\documentclass[11pt,a4paper]{article}

\usepackage[utf8]{inputenc}   
\usepackage[T1]{fontenc}          
\usepackage{graphicx}                   
\usepackage{float}                      
\usepackage{geometry}                   
\usepackage{hyperref}                   
\usepackage{setspace}                   
\usepackage{color}                      
\usepackage[table]{xcolor}              

\hypersetup{                            
        colorlinks=true,                
        breaklinks=true,                
        urlcolor=black,                 
        linkcolor=black,                
        pdftitle={},                    
        pdfauthor={O. Bailleux},        
        pdfproducer={O. Bailleux},      
        pdfsubject={PPC},               
        pdftoolbar=true,                
        pdfmenubar=true,                
        pdfstartview={FitH},            
}

\sffamily                               
\title{\vspace{-1cm}Evolving difficult \textsc{sat} instances thanks to local search}

\normalsize\author{Olivier Bailleux, Université de Bourgogne}

\begin{document}

\sloppy 


\maketitle 



\begin{spacing}{1.05}

\begin{abstract}
We propose to use local search  algorithms to produce \textsc{sat} instances which are harder to solve than randomly generated k-\textsc{cnf} formulae. The first results, obtained with rudimentary search algorithms, show that the approach deserves further study. It could be used as a test of robustness for \textsc{sat} solvers, and could help to investigate how branching heuristics, learning strategies, and other aspects  of solvers impact there robustness.

\end{abstract}

\section{Introduction}

Because the hardest instances for a given solver are not necessarily the hardest ones for another solver, it is difficult to evaluate and to compare the performances of \textsc{sat} solvers.
In the competitions like \cite{sat-comp-2007}, the solvers are evaluated with several kinds of instances (random, hand crafted, industrial...), and the results show that no solver outperforms the other ones in all the categories.

The underlying question is about the \emph{robustness} of \textsc{sat} solvers. How to identify the hardest instances for a given solver ? We propose to use local search  algorithms to find hard instances.

\section{Example \label{section:example}}

In this section, we report the results of a first prospective experimentation using the \textsc{sat4j} framework \cite{sat4j}. The \textsc{sat} solver provided by the \textsc{java} library \textsc{Sat4j} can either be used as a stand alone solver or integrated into a \texttt{java} application. We used this second way to develop a prototype allowing the evolution of a \textsc{sat} instance by local search.

A \textsc{sat} instance is a \textsc{cnf} propositional formula, i.e., a conjunction of clauses, where each clause is a disjunction of literals. Each literal is either a propositional variable or a negated propositional variable. Our initial formula contains $m$ randomly generated clauses of $k$ literals. The variables are uniformly drawn in a set of $n$ variables under the constraint that no variable can occur more than one times in any clause.
Each variable is negated with probability $1/2$.

We propose two ways to evaluate the hardness of an instance, which are the number of propagations and the number of decisions required to solve it. A propagation consists to fix a variable thanks to the unit resolution rule, which is the basic filtering method used in \textsc{sat} solvers.
A decision is a binary node of the search tree.

The used local search process aims to increase the number $p$ of propagations the solver needs to solve the current formula. At each generation, the current instance is modified by replacing an existing randomly drawn clause by a new randomly generated clause.
Then the solver is ran and the new number $q$ of propagations is compared to $p$. If $q<p$ then the old current formula is restored, else the the new one is kept.

\begin{figure}[h]
\begin{center}
\includegraphics[width=10cm]{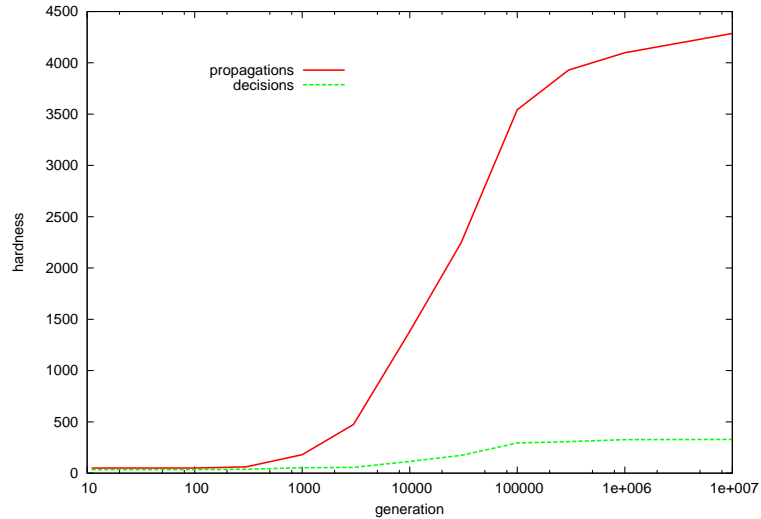}
\caption{Evolving a \textsc{cnf} formula with 100 clauses and 50 variables\label{fig1}}
\end{center}
\end{figure}

Figure \ref{fig1} show the evolution of the number of propagations and the number of decisions during $10^7$ generations. The required number of propagations increases two orders of magnitude, while the required number of decisions increases one order of magnitude. The instance remains satisfiable throughout the evolution.

\section{Evolutions strategies}

The local search algorithm used in section \ref{section:example} is based on a rudimentary greedy search strategy with a transition operator that replaces a randomly drawn clause by a new randomly generated one. Two ways can be explored toward more sophiticated evolution schemes.

\subsection{Search strategies}

The aim of a local search strategy is essentially to avoid the stance of evolution, which may be due to local extremums. We propose three research directions.

\begin{enumerate}
\item 
The algorithm presented in section \ref{section:example} could be improved by modifying several clauses simultaneously when the score (i.e., the number of propagations, the number of decisions, or any other relevant indicator) does not increase for some time. Such a "break" is expected to allow the search process to escape from the current basin of attraction (if applicable), with the hope of finding a more promising evolution path.

\item
Another idea is to try guiding  the search by focusing it on some variables : A weight is assigned to each variable. When changing a clause improve the current score, the weights of the related variables increases. The clauses containing variables with highest weight are preferentially modified. On the contrary, when changing a clause does not improve the current score, the weights of the corresponding variables decreases.

\item
Population based search techniques, like genetic algorithms, could be used. This suppose to find relevant crossover operators, in the sense that the offspring of two (or more) formulae should be likely to be as hard as its parents. An idea worth exploring is first to aggregate two formulae, then reduce the size of the resulting offspring tanks to a  local search process designed to remove clauses. A more standard crossover operator is used in \cite{evolvingsat06} with convincing results.  
\end{enumerate}

Moreover, it is also possible to start the search with a formula already known to be hard with respect to its size.

\subsection{Transition operators}

It is known that the choice of the transition (or neighborhood) operator can dramatically impact the efficiency of a local search process. In section \ref{fig1}, this operator consists to replace a clause by a new one. The number of clauses remains constant during the search. Other ways can be considered, like the following ones.

\begin{enumerate}
\item 
Instead of changing a whole clause, we can try to change one literal at each generation.

\item
Sometimes, a new clause can be added, or an existing one can be removed.

\item
The current instance can be maintained satisfiable or unsatisfiable by rejecting the candidates that do not verify the criterion.
\end{enumerate}

\section{A few experiments}

In this section, we present the first results obtained with a local search algorithm designed to evolve an unsatisfiable formula in two stages. The initial formula is randomly generated with a ratio number of clauses / number of variables which is substantially above the satisfiability threshold \cite{ Dubois2001187}, in such a way that the obtained \textsc{sat} instance is most likely unsatisfiable.
In the first stage, the transition operator consists in removing a clause and the selection criteria accepts the new formula if its remains unsatisfiable. The duration of this stage is 10 times the number of clauses.
In the second stage, the transition operator consists in replacing a randomly chosen clause by a new randomly generated one. The selection criterion accepts the new formula if the number of decisions required to solve it does not decrease.

\begin{figure}[h]
\begin{center}
\includegraphics[width=10cm]{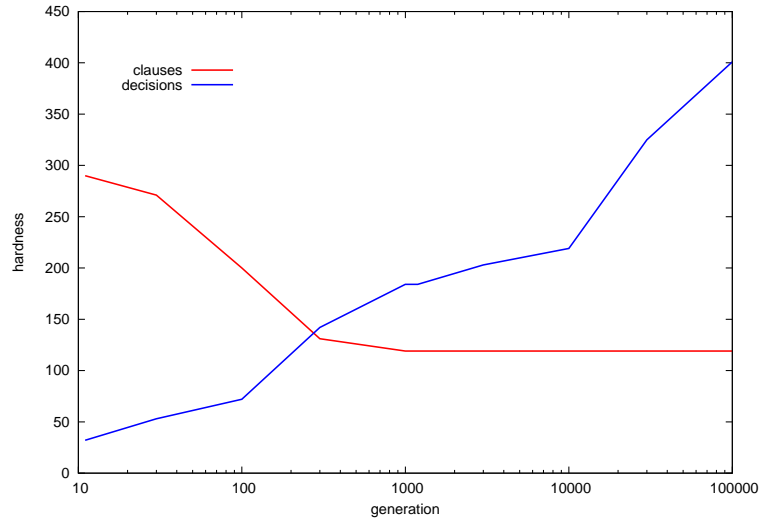}
\caption{Evolving an unsatisfiable \textsc{cnf} formula with 50 clauses and 300 variables\label{fig2}}
\end{center}
\end{figure}

Figure \ref{fig2} shows the evolution of both the size of the $3$-\textsc{cnf} formula of 50 variables and the number of decision required to solve it on $10^5$ generations. The second stage begins after about 1000 generations.

\begin{figure}[h]
\begin{center}
\includegraphics[width=10cm]{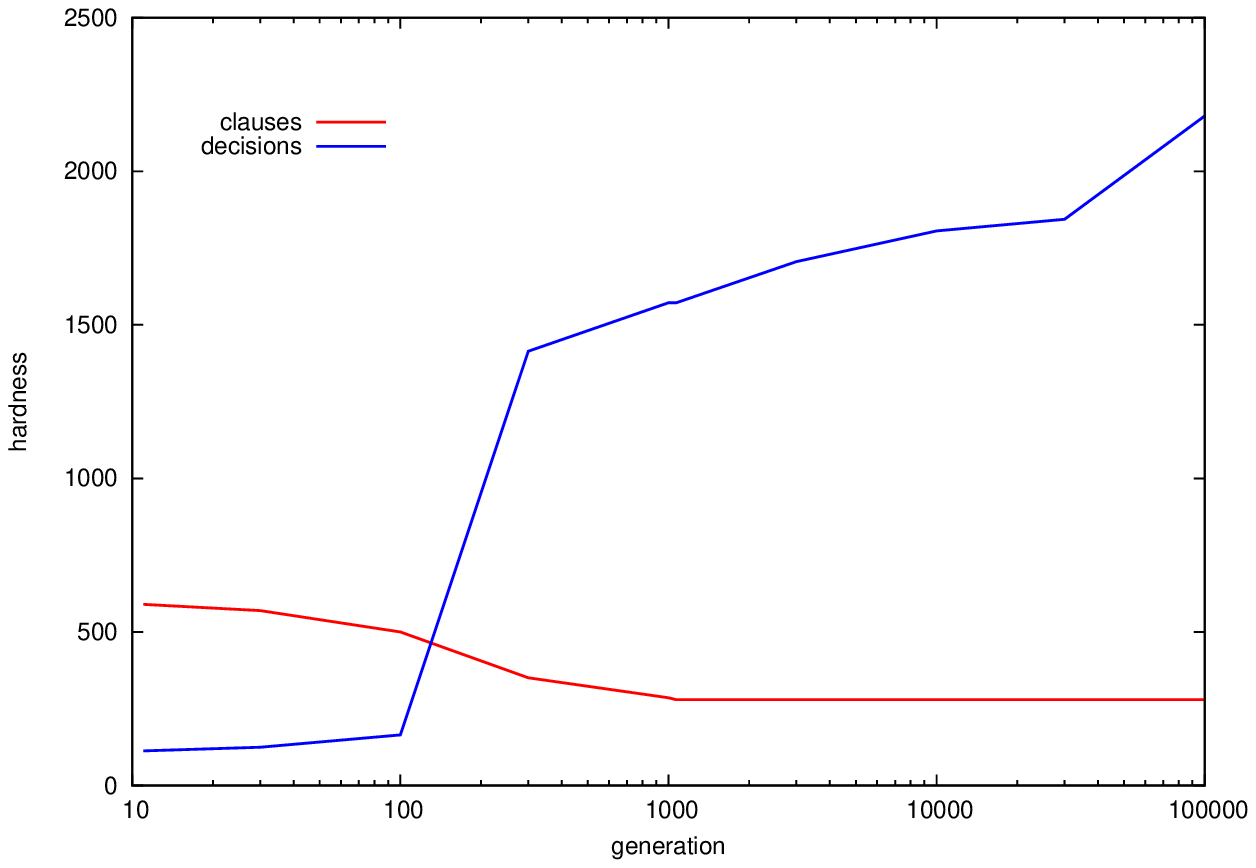}
\caption{Evolving an unsatisfiable \textsc{cnf} formula with 100 clauses and 600 variables\label{fig3}}
\end{center}
\end{figure}  

Figure \ref{fig3} shows the evolution of a $3$-\textsc{cnf} formula of 100 variables in the same conditions. The number of decisions increases from 122 to 1572 during the first stage, where the number of clauses decreases from 600 to 280. In the second stage, the number of clauses remains constant while the numbers of decisions increase up to 2180.
We examined whether the hardness of this instance also increases with respect to another \textsc{sat} solver, i.e., cryptominisat \cite{cryptominisat} : the initial formula requires 91 decisions, and the final one 2051 decisions.

\begin{figure}[h]
\begin{center}
\includegraphics[width=10cm]{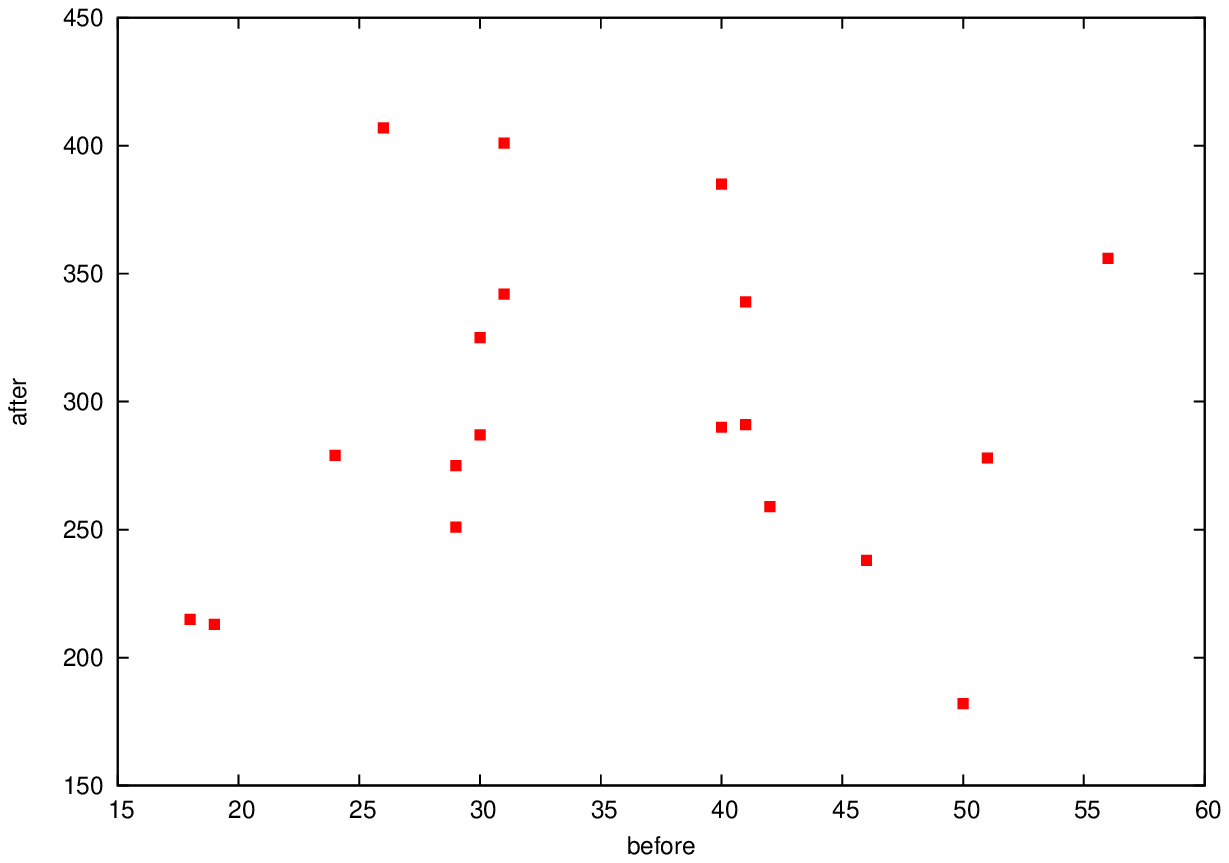}
\caption{Evolving 20 unsatisfiable \textsc{cnf} formula with 50 clauses and 300 variables, initial versus final score\label{fig4}}
\end{center}
\end{figure}

For a first glimpse of how the initial instance impacts the result of the evolution, Figure \ref{fig4} shows the initial and final number of decisions for twenty formula of 50 variables : there is no apparent correlation between the initial and final hardness, which is consistently greatly increased.

\section{Related works}

We presented the idea of using local search for finding hard \textsc{sat} instances in \cite{Bailleux96b}, but this work, although promising, has not been resumed since.

To the best of our knowledge, the only other work really close to our approach is \cite{evolvingsat06}, which use genetic algorithms to evolve combinatorial problem instances in order that there become difficult to solve. This already fairly advanced work proposes an analysis of the generated instances for identifying what make them hard to solve.

\section{Concluding remarks}

We proposed to use local search  algorithms  to produce \textsc{sat} instances which are harder to solve than randomly generated k-\textsc{cnf} formulae.

The specificity of the proposed approach is that the computation of the objective function is very expensive, because the aim is to produce hard instances, with the result to increase the cost of evaluate the hardness of these instances.

Nevertheless, the first results obtained with rudimentary search algorithms show that the approach deserves further study. It could be used as a test of robustness for \textsc{sat} solvers, and could help to investigate how branching heuristics, learning strategies, and other criteria, impact
this robustness.

\bibliographystyle{plain}
\bibliography{evol-diff} 

\end{spacing}

\end{document}